\documentclass[runningheads]{llncs}
\usepackage{graphicx}
\usepackage{subfigure}
\usepackage{algorithm}
\usepackage{algorithmic}
\usepackage{enumitem}
\usepackage{graphicx}
\usepackage{booktabs}

\begin{document}
\title{Controlled AutoEncoders to Generate Faces from Voices}

\author{Hao Liang\inst{1} \and
Lulan Yu\inst{1} \and
Guikang Xu\inst{1} \and
Bhiksha Raj\inst{1} \and
Rita Singh\inst{1}}
\authorrunning{Liang. et al.}
\institute{Carnegie Mellon University, Pittsburgh PA 15213 USA\\
\email{hliang2@andrew.cmu.edu, lulany@alumni.cmu.edu,
guikangx@alumni.cmu.edu,
\{bhiksha, rsingh\}@cs.cmu.edu}}

\maketitle              % typeset the header of the contribution

\begin{abstract}
    Multiple studies in the past have shown that there is a strong
correlation between human vocal characteristics and facial features. 
However, existing approaches generate faces simply from voice, without exploring the set of features that contribute to these observed correlations. 
A computational methodology to explore this can be devised by
rephrasing the question to: ``how much would a target face have to change in order to be perceived as the originator of a source voice?''
With this in perspective, we propose a framework to morph a target face in response to a given voice in a way that facial features are implicitly guided by learned voice-face correlation in this paper. 
Our framework includes a guided autoencoder that converts one face to another, controlled by a unique model-conditioning component called a \textit{gating controller} which modifies the reconstructed face based on input voice recordings.  
We evaluate the framework on VoxCelab and VGGFace datasets through human subjects and face retrieval.
Various experiments demonstrate the effectiveness of our proposed model. 

\end{abstract}
\noindent\textbf{Index Terms}: controlled autoencoder, face generation, speech-face association learning, voice-face biometric matching

\section{Introduction}

Curiosity has long driven humans to picture what someone else looks like upon hearing his/her voice. One of the assumptions, presumably made based on millennia of experience, is that there exists a statistical relation between what a person {\em looks} like, and what they {\em sound} like \cite{article}. 
Some of these relations are immediately obvious, {\em e.g.} we expect the visage that goes with old-sounding voices to look old, and that of female voices to be female. 
Besides these obvious covariates, however, it may be expected that there are other unapparent dependencies between human vocal characteristics and facial features, since humans are able to associate faces with voices with surprising accuracy \cite{Kim2018on}.

In trying to visualize the face behind a voice, however, we do not generally begin {\em de novo}, drawing up the face from whole cloth. 
Indeed, if one were to try to imagine the source of a sound such as voice with no prior information at all, it is highly unlikely that one would picture anything resembling a face. 
Therefore, we visualize faces based on two hypotheses:
\begin{itemize}[leftmargin=*]
    \item First, that in identifying the sound as a human voice, we draw upon the general shape of human faces, rather than any physical object that could produce the sound, and 
    \item Second, that we draw upon an actual imagined face and make the modifications we deem necessary to match the voice.
\end{itemize}

The problem of {\em automatically} guessing speakers' faces from voice recordings has recently gained some popularity (or notereity, depending on the reader's perspective) \cite{Oh_2019_CVPR,DBLP:journals/corr/abs-1806-00154,Wiles18}. The proposed approaches have taken two paths.  The first, ``matching'' approach  finds the most suitable face from a collection \cite{wen2018disjoint,Nagrani18a}, which is fundamentally restricted by the selection of faces it can choose from and cannot generate entirely novel faces. The second, ``generation'' approach attempts to {\em synthesize} the facial image \cite{Oh_2019_CVPR}. The approach embodies the first of our two hypotheses above : the proposed solutions learn a generic statistical model for the relationship between voices and face-like images, which they use to generate faces from novel voices. However, learning such statistical models is fraught with challenge. As \cite{wav2pix2019icassp} discover, direct modelling of statistical dependencies can result in models that often do not generate images that are even facelike. More sophisticated approaches do manage to generate face-like imagery, but these often show distortions \cite{2018arXiv180807276L,wen2019face}. While one may nevertheless try to produce better-quality visualizations, these techniques are dependent more on the well-known ability of GANs and similar models to hallucinate near-perfect faces \cite{PMID:31589460,wen2019face}, than on any dependence of these faces on all but the highest-level aspects of the voice, such as gender and age. Even so, they will still suffer problems such as {\em mode collapse} (whereby all voices end up producing the same face)\cite{Oh_2019_CVPR,8852154}, and lack of variety in the generated faces \cite{PMID:31589460}.

In this paper we take a different approach based on the second of our hypotheses above: that it may be easier to generate realistic facsimiles of a speaker's face if one were to begin with an initial template of an actual face and only make the minimal changes necessary to it (retaining as many features from the input face as possible), to make it match the voice. Accordingly, we propose a new model that begins with an initial ``proposal'' face image, and makes adjustments to it according to information extracted from the voice.

Our proposed model, which we will refer to as a {\em controlled AutoEncoder} (CAE), comprises two components. The primary component is a  scaffolding  based on a  U-net-like~\cite{inproceedings} autoencoder, which takes in the proposal face and outputs a redrawn face. In the absence of other signals, the network must regenerate the input proposal face. The second component, which accepts the voice input, is a gating {\em controller} that modifies the parameters of the decoder based on information derived from the voice, to modify the generated face to match the voice. While being trained, the model also includes a discriminative component that is used to optimize the parameters of the network. The entire model is learned using a set of losses that simultaneously capture the degree of match to the target (speaker's) face and deviation from the proposal face. The CAE has several advantages over prior models. Being fundamentally an autoencoder, the output does not diverge from a face. As a morphing mechanism, rather than a genreating one, it does not face problems such as mode collapse. Diversity of output may be obtained simply by changing the proposal face.

Experimental results show that our model performs well under several evaluation metrics, including human subjective evaluation, feature similarity and face retrieval performance. 

\section{The controlled AutoEncoder model}

The purpose of our model is to modify an input ``proposal'' face image to best match a given voice. The model has two components:  an autoencoder based on a U-net-like~\cite{inproceedings} structure that derives a latent representation of the proposal face and attempts to reconstruct the latter from it, and a gating {\em controller} that modifies the parameters of the decoder based on the input speech signal. We describe the two below.

\subsection{The AutoEncoder scaffold}

The main scaffold of our model is an autoencoder that derives a latent representation from the proposal face and reconstructs it from this representation.  We specifically choose the  U-net structure~\cite{inproceedings} for reasons we will shortly explain.  The U-net includes an {\em encoder} that comprises a sequence of downsampling convolutional layers, ending finally in a simple, low-dimensional representation of the input image. This is followed by a {\em decoder} that comprises a sequence of upsampling {\em transpose-convolutional} layers that recreate the input image from the latent representation. The structure of the entire network is symmetric around its central latent-representation layer --  corresponding to each  layer of the encoder is a complementary decoder layer that is at an identical distance (or depth) from the central latent-representation layer, and has the same size as its encoder complement. 

What distinguishes the U-net from conventional autoencoders is that the U-net also includes {\em direct} ``skip'' connections from each encoder layer to its complementary decoder layer. This is particularly relevant to our model, since it provides a pathway for information about the details of the face to be retained, while permitting high-level latent characteristics to be modified.

\begin{figure}[ht]

\centering

\includegraphics[scale=0.15]{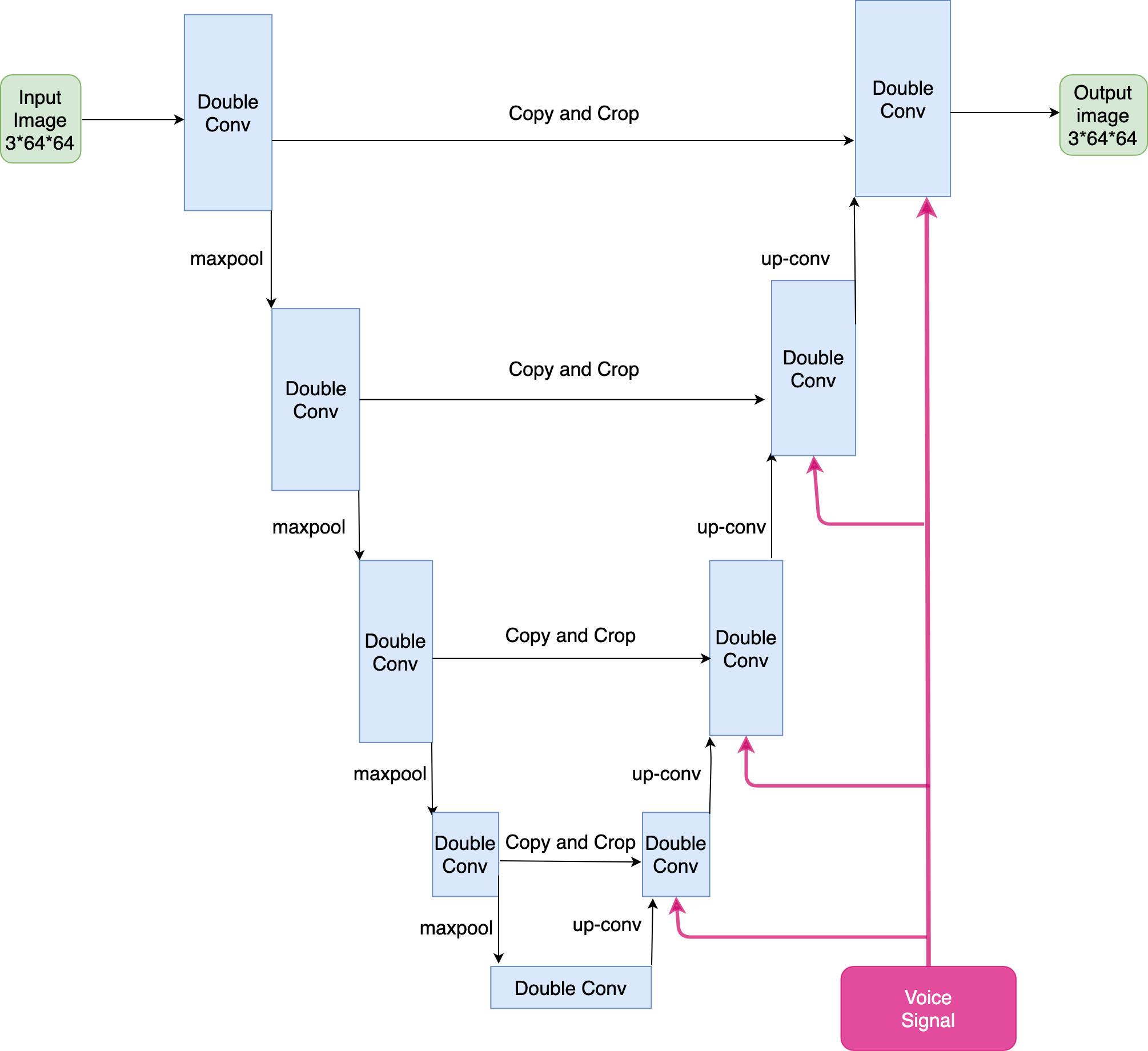}      

\caption{\label{fig3.1}\textbf{The controlled AE model architecture}. DoubelConv represents a layer with two convolutional neural networks, with a batchnorm layer appending to each of them.}

\end{figure}

The blue components of Figure \ref{fig3.1} illustrate the U-net AE scaffolding of our model.

In our implementation the encoder consists of 4 convolutional layers with 64,128,256 and 512 output channels. All convolutional filters are size $3 \times 3$. Convolutional layers followed by maxpooling layers of size $2 \times 2$, with stride 2. The decoder is the mirror image of the encoder, sans the pooling layers. The transpose convolution in the decoder has a stride of size 2. Input images are rescaled to $3\times 64 \times 64$ and output images too are the same size.

\subsection{The gating controller}

The gating controller modifies the {\em trasverse convolutional filters} of the decoder, based on the voice signal.

In our setup, the voice recordings, duly endpointed, are first transformed into a 64-dimensional log mel-spectrographic representation, computed using an analysis window of 25ms and a frameshift of 10ms. The resulting spectrographic representation is fed to an {\em embedding} network, which produces a 64-dimensional embedding of the voice signal. The voice embedding is subsequently input, in parallel, to a number of linear layers, one each corresponding to each layer of the decoder.  The linear layer corresponding to each decoder layer produces as many outputs as the number of convolutional filter parameters in the decoder layer. The outputs of the linear layer are passed through a sigmoid activation. The outputs of the sigmoid multiply the parameters of the filters, thereby modifying them.

\begin{figure}[ht]
 
\centering
\includegraphics[scale=0.25]{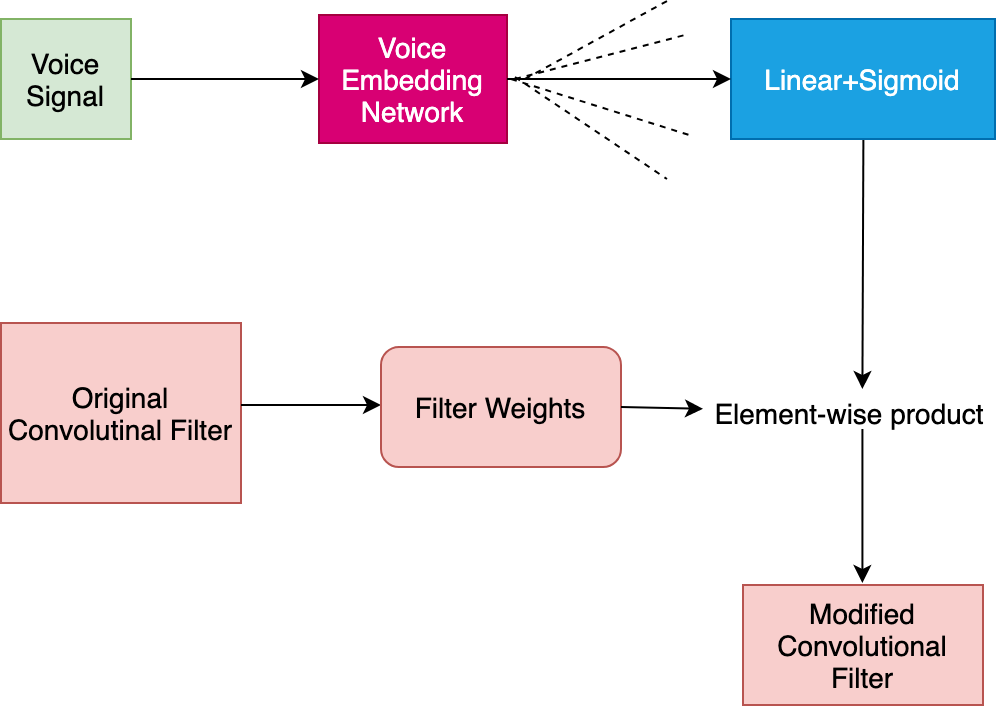}        
\caption{The gating controller}
\label{fig3.2}
 
\end{figure}

The pink components of Figure \ref{fig3.1} illustrate the overall operation of the gating controller. 
%Algorithm 1 explains the gating algorithm. 
Figure \ref{fig3.2} shows the details of the gating control for {\em one} of the decoder layers. The dotted lines in the figure show paths to other layers.

The embedding network is a key component of this framework, as it is intended to extract speaker-specific information from the voice. For our work we employ an architecture derived from  \cite{wen2019face}, where it was found to be effective for biometric voice-to-face matching. Table \ref{tab:network} shows the architecture of the embedding network. It too is a convolutional network $t_0$ $t_1$ {\em etc.} are the lengths (temporal) of the the input after each layer, which compresses the input length with a stride of 2. We found it most effective to pre-train the voice embedding network, using the DimNet framework from  \cite{wen2019face}, and to subsequently fix the network.

\begin{table}
\centering
\caption{\textbf{Detailed network architecture.} In voice embedding network, Conv $3_{/2,1}$ denotes a 1D convolutional layer with kernel size 3, stride 2 and padding 1. BN denotes batchnorm. In discriminator/classifier, Conv $1 \times 1_{/1,1}$ denotes a 2D convolutional layer with kernel size $1 \times 1$, stride 1 and padding 0. They share these structures, and for the last layer, discriminator has: FC $64 \times 1$, Sigmoid as activation, and output shape is 1. Classifier has: FC $64 \times k$, Softmax as activation, and output shape is k.}
% % % \vspace{-1em}
\begin{tabular}{c|c|c}  
\toprule
\multicolumn{3}{c}{Voice Embedding Network} \\
\hline
Layer  & Act. & Output shape \\
\midrule
Input       & -  & $64 \times t_0$      \\
Conv $3_{/2,1}$            & BN+ReLU    & $256 \times t_1$      \\
Conv $3_{/2,1}$            & BN+ReLU    & $384 \times t_2$      \\
Conv $3_{/2,1}$            & BN+ReLU    & $576 \times t_3$      \\
Conv $3_{/2,1}$            & BN+ReLU    & $864 \times t_4$      \\
Conv $3_{/2,1}$            & BN+ReLU    & $64 \times t_5$      \\
AvgPool $3_{/2,1}$            & -   & $64 \times 1$      \\

\bottomrule
\end{tabular}
\begin{tabular}{c|c|c}  
\toprule
\multicolumn{3}{c}{Discriminator/Classifier} \\
\hline
Layer  & Act. & Output shape \\
\midrule
Input       & -  & $3 \times 64 \times 64$      \\
Conv $1 \times 1_{/1,0}$   & LReLU    & $32 \times 64 \times 64$      \\
Conv $3 \times 3_{/2,1}$ .  & LReLU    & $64 \times 32 \times 32$      \\
Conv $3 \times 3_{/2,1}$    & LReLU    & $128 \times 16 \times 16$       \\
Conv $3 \times 3_{/2,1}$   & LReLU    & $256 \times 8 \times 8$      \\
Conv $3 \times 3_{/2,1}$   & LReLU    & $512 \times 4 \times 4$       \\
Conv $4 \times 4_{/1,0}$   & LReLU    & $64 \times 1 \times 1$      \\
% FC $64 \times 1$/    FC $64 \times k$        &-   & 1/k     \\

\bottomrule
\end{tabular}
\label{tab:network}

\end{table}

\section{Training and inference}

\subsection{Training the network}

The network is trained on a collection of training instances, where each instance comprises an original face, a target voice and the corresponding target face.
We optimize the parameters of the network to try to ensure that the output of the U-net stays close to its input, while simultaneously capturing the identity of the speaker. To support this, during the learning phase, we include two additional components, that work on the image produced by the model:
\begin{itemize}[leftmargin=*]
    \item A binary {\em discriminator} that is intended to ensure that the generated images are facelike. The architecture for which is also described in Table \ref{tab:network}.
    \item A multi-class {\em classifier} which tries to enforce the visual identity of the speaker on the generated face. The classifier is a shared structure with the discriminator, with the exception that the final representation layer of the discriminator forks off into a multi-class softmax layer with as many outputs as we have subjects in the training data.
\end{itemize}

We employ a number of losses to train the network. We explain these below. In the explanation we represent real faces with $\textit{f}$, synthetic faces with $\textit{$\hat{f}$}$, voice with $\textit{v}$, and identity label with $\textit{I}$. In terms of model components, we use $G$ to represent the generator (the scaffolding), $D$ for the discriminator, and $C$ for classifier.
\begin{itemize}[leftmargin=*]
    \item \textbf{L1-norm loss}: $\textit{L}_{L1}$(\textit{$f_1$},\textit{$f_2$}) computes the total $L_1$ norm of the error between the RGB values of two face images $f_1$ and $f_2$. 
    \item \textbf{Classifier loss}: $\textit{L}_c$(\textit{f},\textit{I}) is the cross-entropy between the classifier's output and a one-hot representation of the identity of the speaker.
    \item \textbf{Discriminator loss}: $\textit{L}_d$(\textit{f},\textit{I}) computes a binary cross entropy between the discriminator output and a binary label that specifies if the image input to the discriminator is synthetic or real.
\end{itemize}

The actual training paradigm, which is inspired by \cite{wen2018disjoint,wen2019face}, StarGAN~\cite{choi2018stargan} and CycleGAN~\cite{8237506}, tries to simultaneously satisfy multiple objectives. 

First, the output of the generator must be similar to the target face, \textit{i.e.} $G(f_a, v_b) \approx f_b$, where $f_a$ denotes the face image of original identity $A$ and $v_b$ denotes the voice embedding of target identity $B$. In addition, the generated face should resemble the original identity, \textit{i.e.} $G(f_a, v_b) \approx f_a$. Likewise, the generator should be capable of generating the original face back when taking the fake image $B$ and voice of $A$ as input, \textit{i.e.} $G(G(f_a, v_b), v_a) = f_a$. The functionality of the discriminator is determining if an input image is synthetic or natural, \textit{i.e.} $D(f) = 1, D(\hat{f}) = 0$. The goal of the classifier is to determine if the image produced in response to a voice is indeed the face of the target, i.e. is $C(\hat{f_B)} = B$. 
With the aforementioned building blocks,  the complete training process as is explained in Algorithm 1.
\begin{algorithm}[tb]
\caption{Training algorithm of Our model}
\label{alg:algorithm}
\textbf{Input}: A set of voice recordings and labels(V, $I_v$). A set of labeled face images with labels(F, $I_f$). A pre-trained fixed voice embedding network E(v) tuned under a speaker recognition task.\\
% \textbf{Parameter}: Optional list of parameters\\
\textbf{Output}: Generator parameters $\theta_g$
\begin{algorithmic}[1] %[1] enables line numbers.
\WHILE{not converge}
\STATE Randomly sample a face image from F $(f_A, I_A)$;
\STATE Randomly sample a voice recording from V $(v_B, I_B)$;
\STATE Retrieve the corresponding face image of B $(f_B, I_B)$;
\STATE Retrieve the corresponding voice recording of A $(v_A, I_A)$;
\STATE Get voice embedding $e_B$ = E$(v_B)$;
\STATE Update the discriminator D(f; $\theta_d$) with loss\\
    \qquad $\Delta_{\theta_d}$($L_d$(G($f_A$,$e_B$;$\theta_d$), 0) + $L_d$($f_A$,1));
\STATE Update the classifier C(f; $\theta_c$) with loss\\
\qquad $\Delta_{\theta_c}$( $L_c$(C($f_A$;$\theta_c$),$I_A$));
\STATE Get fake face of B $\hat{f_B}$ = G($f_A$,$e_B$;$\theta_g$);
\STATE Get voice embedding $e_A$ = E$(v_A)$;
\STATE Update the generator with loss\\
$\Delta_{\theta_g}$($\lambda_1$$L_{L1}$($\hat{f_B}$,$f_A$) + $\lambda_2$$L_{L1}$($\hat{f_B}$,$f_B$) + $\lambda_3$$L_{c}$(C($\hat{f_B}$;$\theta_c$),$I_B$) + $\lambda_4$$L_d$(D($\hat{f_B}$;$\theta_d$),1) + $\lambda_5$$L_{L1}$(G($\hat{f_b}$,$e_A$; $\theta_g$),$f_A$));
\ENDWHILE
\STATE \textbf{return} $\theta_g$
\end{algorithmic}
\end{algorithm}

% \vspace{-1em}
\subsection{Inference}
% \vspace{-0.5em}
During the inference time, a proposal face image $f_a$ and a target voice's embedding $v_b$ are injected to the generator network, and a face image $\hat{f}$ is synthesized.
% \vspace{-1em}
\section{Experiments}
% \vspace{-0.5em}
\subsection{Dataset}
% \vspace{-0.5em}
In this paper, we chose VoxCeleb\cite{Nagrani17} and VGGFace\cite{parkhi2015deep} datasets, given that there exists intersection on the celebrity identities presented in the two datasets. Specifically, 149,354 voice recordings and 139,572 face images of 1,225 subjects are found in the intersection in total.
% \vspace{-1em}
\subsection{Experiments}
\begin{figure*}[!hbtp]
    \centering

    \subfigure[]{\label{fig:a}\includegraphics[height=36mm]{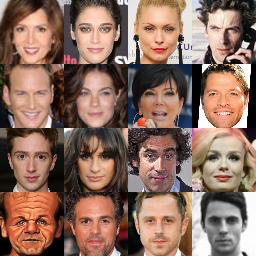}}
     \hspace{8 mm} 
    \subfigure[]{\label{fig:b}\includegraphics[height=36mm]{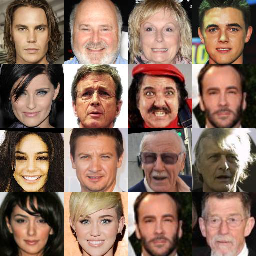}}

    ~
    
    \subfigure[]{\label{fig:c}\includegraphics[height=36mm]{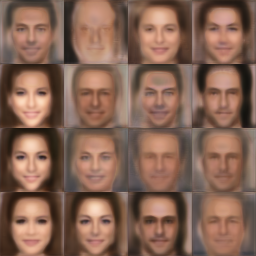}}
     \hspace{8 mm} 
    \subfigure[]{\label{fig:d}\includegraphics[height=36mm]{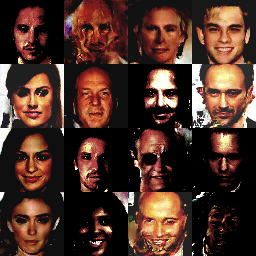}}
    
    \subfigure[]{\label{fig:e}\includegraphics[height=30mm]{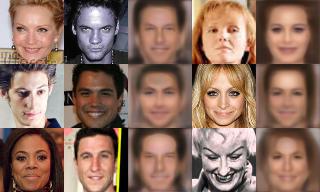}}

    \subfigure[]{\label{fig:f}\includegraphics[height=30mm]{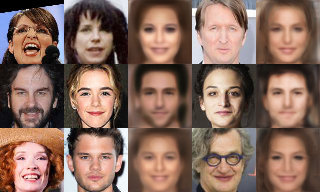}}

    \caption{\textbf{Experiment results} (a) Original face (b) Target face corresponding to input voice (c) Generated face using CAE (d) Generated face using \cite{wen2019face} (e) the first column shows the proposal face. subsequent pairs show target and generated faces. (f) The first column shows the target face. Subsequent pairs show proposal and generated faces. (Best viewed in color).}
    \label{fig:res2}

\end{figure*}

% \vspace{-0.5em}
We used Adam optimizer with batch size 1. Note that it makes no sense using a value of batch size other than 1. The reason being, in our proposed framework, each pair of voice and face is unique, thus it requires a unique set of network weights to perform generation. After grid search, we finalized the learning rate to be 0.0002, $\beta_1$ = 0.5, and $\beta_2$ = 0.999. Also, we trained the discriminator and generator with ratio 1:1, using $\lambda_1$ = 1, $\lambda_2$ = 10, $\lambda_3$ = 1, $\lambda_4$ = 1, $\lambda_5 = 10$. Results for this part are illustrated in Figure~\ref{fig:res2}.

To validate the the model indeed performs as we expect, we present visualizations of the following settings:
\begin{itemize}[leftmargin=*]
    \item \textbf{Fix the proposal face, change the input voice.} This is to confirm our model's ability of generating different faces based on the same proposal face. 
    \item\textbf{Fix the input voice, change the proposal face.} This is to confirm our model's ability of generating the target face based on different proposal faces. 
\end{itemize}
Example results are shown in Figures~\ref{fig:e} and \ref{fig:f} respectively.
Results are random selections from our data and are not cherry picked. We can confirm that our model is able to generate a target face by performing necessary modifications to the proposal faces. Figures ~\ref{fig:a}, ~\ref{fig:b}, ~\ref{fig:c}, ~\ref{fig:d} also compare faces generated from voices by our CAE to those generated by the technique from \cite{wen2019face}, which is currently a state of the art. For example, the faces in the upper left corner of (a), (b), (c), (d) represent the original input face, corresponding face of the target voice, generated face using our model, generated face using \cite{wen2019face} respectively.

\begin{figure*}[!hbtp]
    \centering
    
    {\label{fig:g}\includegraphics[height=30mm]{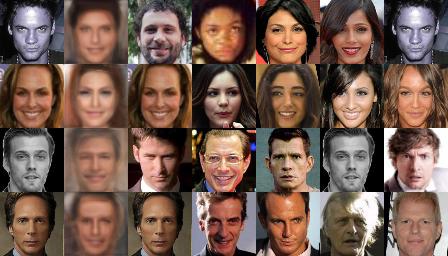}}
    % \vspace{-1em}
    \caption{\textbf{Face retrieval results}. The first column corresponds to the target face, the second column corresponds to the generated face, and the following five columns correspond to the retrieval results, in ranked order. (Best viewed in color).}
    \label{fig:res3}
% \vspace{-1.5em}
\end{figure*}

\subsection{Evaluation}
% \vspace{-0.5em}
Evaluation of generative models like GANs models is always tricky and several methods have been proposed ~\cite{BORJI201941} for it. Considering the speciality of our task which requires the output image to keep features from both faces, however, classification error ~\cite{wen2019face}, Fréchet Inception Distance (FID) ~\cite{NIPS2017_7240} and reconstruction error, which are the most common evaluation metrics, are not good options. Instead, we resort to  human subjective votes -- one of the most mature and solid method -- as our first evaluation metric, and {\em retrieval} as our second.
% \vspace{-1em}
\subsubsection{Human subjective evaluation}
% \vspace{-0.5em}
The human subject method is broadly adopted in many generative models~\cite{DBLP:conf/nips/ZhouGKNFB19,bau2019gandissect,10.5555/3294996.3295075,10.5555/3298483.3298649} because of its intuitiveness and reliability when the number of samples becomes large. In this task, we conducted human subjective evaluation through interviews. To ensure the feasibility and reliability, the interviewers were first asked to finish a  real-image mapping, which contains two images and their corresponding voices, and map the images with the voices. A 89\% mapping accuracy was obtained from 20 subjects, performing 100 tests each,  validating our conjecture: human subjective mapping between voices and images is reliable and robust. Next, they were shown a generated face image that was synthesized with our model, its corresponding voice audio, and another randomly generated face image which has the same gender, then asked to map the voice to one of the two images. 

All the interviewers were asked to rate in total 100 rounds of tests. Finally, a 72\% mapping accuracy was achieved. Based on the result, we are convinced we have reached our goal -- to generate a face behind a voice by only modifying necessary features from a proposal face.

\begin{table}
\centering
\caption{\textbf{Human subjective evaluation accuracy.} Random guess is set up as a baseline, which is 50\%.}
\begin{tabular}{lr}  
\toprule
Task & Accuracy (\%) \\
\midrule
real image maping       & 89     \\
generated image mapping(CAE)  & \textbf{72}      \\
% generated image mapping(baseline)  &  59     \\
\hline
random                   & 50 \\
\bottomrule
\end{tabular}
\label{tab:acc}
% \vspace{-1em}
\end{table}

% \vspace{-1em}
\subsubsection{Feature similarity and face retrieval}
% \vspace{-0.5em}
To make our results more convincing, we adopted the evaluation strategies from ~\cite{Oh_2019_CVPR}. We reused the face embedding network we got from our training to get three 64-dimensional feature vectors for the generated face images, original input image and the corresponding image of the voice respectively. Then we calculated the cosine similarity between the generated image's embedding vector and the other two. The result is shown in table~\ref{tab:cos}, indicating out results complete the goal of maintaining features from both faces.

\begin{table}
\centering
\caption{\textbf{Feature cosine similarity.} Here $g_C$, A, B denotes generated face image from our model(CAE), input image and corresponding face image of input voice respectively.}
\label{tab:cos}
\begin{tabular}{lr}  
\toprule
Mapping pair & cos \\
\midrule
$g_C, A$       & \textbf{0.32}     \\
$g_C, B$       & \textbf{0.42}       \\
% $g_B, A$       & 0.28     \\
% $g_B, B$       & 0.32     \\
\hline
random mapping  & 0.21 \\
\bottomrule
\end{tabular}
% \vspace{-2em}
\end{table}
% % \vspace{-2em}
In addition to this, we also utilized face retrieval, where we reused the classifier we obtained from  training again and injected the generated face image as input, and got top 5 faces retrieved, i.e. whose corresponding value in the output vector of classifier lies in top 5. If the original face was in the top 5, we marked it as a successful retrieval. The successful ratios of our model was \textbf{21\%}, in a 500 images retrieval task. Some samples are shown in Figure~\ref{fig:res3}.

\section{Conclusion}
% \vspace{-0.5em}
We presented a novel model, the Controlled AutoEncoder, for modifying human face images to match input voices. The model uses conditioning voice input as a {\em controller}, to modify output images. While this is seen to work well in the setting of conditional production of face images, we believe the model has wider applicability in other problems, which we are investigating. Within the proposed framework too, much room remains for improvement. Current results seem to indicate that the entire face is modified, as opposed to specific components of it. We are conducting investigations into whether finer control may be exercised over this process.

\bibliographystyle{IEEEtran}
\bibliography{mybib}
\end{document}